%% file: paper.tex
\pdfoutput=1

\documentclass[11pt]{article}

\usepackage[hidelinks]{hyperref}

\input{math.tex}

\usepackage{style}
\usepackage{EMNLP2022}

\usepackage[T1]{fontenc}

\usepackage[utf8]{inputenc}

\usepackage{microtype}

\usepackage{inconsolata}

\usepackage{xspace}
\usepackage{booktabs}
\usepackage{scalerel}

\definecolor{orange}{rgb}{1,0.5,0}
\definecolor{mdred}{rgb}{0.7,0,0}
\definecolor{mdgreen}{rgb}{0.05,0.6,0.05}
\definecolor{mdblue}{rgb}{0,0,0.7}
\definecolor{dkblue}{rgb}{0,0,0.5}
\definecolor{ltblue}{rgb}{0.40,0.40,0.80}
\definecolor{dkgray}{rgb}{0.3,0.3,0.3}
\definecolor{slate}{rgb}{0.25,0.25,0.4}
\definecolor{gray}{rgb}{0.5,0.5,0.5}
\definecolor{ltgray}{rgb}{0.7,0.7,0.7}
\definecolor{purple}{rgb}{0.7,0,1.0}
\definecolor{lavender}{rgb}{0.65,0.55,1.0}

\newcommand{\papercomment}[3]{\ensuretext{\textcolor{#3}{[#1 #2]}}}

\newif\ifcomments
\ifcomments
    \newcommand{\zfw}[1]{\papercomment{\marker{Z}{W}}{#1}{mdgreen}}
    \newcommand{\rob}[1]{\papercomment{\marker{R}{L}}{#1}{red}}
    \newcommand{\sameer}[1]{\papercomment{\marker{S}{S}}{#1}{purple}}
    \newcommand{\ib}[1]{\papercomment{\marker{I}{B}}{#1}{brown}}
    \newcommand{\dirk}[1]{\papercomment{\marker{D}{G}}{#1}{red}}
    \newcommand{\tocite}[1]{\papercomment{\marker{To}{Cite}}{#1}{ltblue}}
    \newcommand{\todo}[1]{\papercomment{\marker{To}{Do}}{#1}{ltblue}}
\else
    \newcommand{\zfw}[1]{}%
    \newcommand{\rob}[1]{}%
    \newcommand{\sameer}[1]{}%
    \newcommand{\ib}[1]{}%
    \newcommand{\dirk}[1]{}%
    \newcommand{\tocite}[1]{}%
    \newcommand{\todo}[1]{}%
\fi

\definecolor{revcolor}{rgb}{0.05,0.8,0.05}
\newcommand{\rev}[1]{\textcolor{black}{#1}}

\definecolor{noprompt}{rgb}{0.7,0.7,0.7}
\newcommand{\mFF}{\textbf{%
    MTL-T%
    \includegraphics[height=\fontcharht\font`\B]{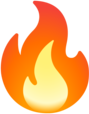}%
    \textcolor{noprompt}{P\xmark}%
}\xspace}
\newcommand{\mFT}{\textbf{%
    MTL-T%
    \includegraphics[height=\fontcharht\font`\B]{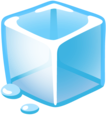}%
    P%
    \includegraphics[height=\fontcharht\font`\B]{fig/fire.png}%
}\xspace}
\newcommand{\mTT}{\textbf{%
    MTL-T%
    \includegraphics[height=\fontcharht\font`\B]{fig/fire.png}%
    P%
    \includegraphics[height=\fontcharht\font`\B]{fig/fire.png}%
}\xspace}

\usepackage{wasysym}
\usepackage{marvosym}

\title{Continued Pretraining for Better Zero- and Few-Shot Promptability}

\author{Zhaofeng Wu$^\text{\Cancer}$ \quad
    Robert L. Logan IV$^\text{\Aries}$ \vspace{-0.18cm} \\ \bf
    Pete Walsh$^\text{\Libra}$ \quad
    Akshita Bhagia$^\text{\Libra}$ \quad
    Dirk Groeneveld$^\text{\Libra}$ \quad
    Sameer Singh$^{\text{\Libra} \rotatebox[y=0.18cm]{180}{\textsuperscript{\Aries}}}$ \quad
    Iz Beltagy$^\text{\Libra}$ \\
    $^\text{\Cancer}$MIT \quad $^\text{\Aries}$Dataminr Inc. \vspace{-0.18cm} \\
    $^\text{\Libra}$Allen Institute for Artificial Intelligence \quad $^{\rotatebox[y=0.18cm]{180}{\textsuperscript{\Aries}}}$University of California, Irvine \\
    \texttt{zfw@csail.mit.edu} \quad \texttt{rlogan@dataminr.com} \\
    \texttt{\{petew,akshitab,dirkg,beltagy\}@allenai.org} \quad \texttt{sameer@uci.edu}
}

\begin{document}
\maketitle
\blfootnote{This work was done when Zhaofeng Wu was at AI2, and Robert Logan was at UCI.}
\blfootnote{We release our code and models at \url{https://github.com/allenai/better-promptability}.}
\begin{abstract}

Recently introduced language model prompting methods can achieve high accuracy in zero- and few-shot settings while requiring few to no learned task-specific parameters.
Nevertheless, these methods still often trail behind full model finetuning.
In this work, we investigate if a dedicated continued pretraining stage could improve ``promptability'', i.e., zero-shot performance with natural language prompts or few-shot performance with prompt tuning.
We reveal settings where existing continued pretraining methods lack promptability.
We also identify current methodological gaps, which we fill with thorough large-scale experiments.
We demonstrate that a simple recipe, continued pretraining that incorporates a trainable prompt during multi-task learning, leads to
improved promptability in both zero- and few-shot settings compared to existing methods, up to 31\% relative.
On the other hand, we find that continued pretraining using MAML-style meta-learning, a method that directly optimizes few-shot promptability, yields subpar performance.
We validate our findings with two prompt tuning methods, and, based on our results, we provide concrete recommendations to optimize promptability for different use cases.

\end{abstract}

\input{sections/10_intro}
\input{sections/20_prompting}
\input{sections/30_improving_promptability}
\input{sections/40_setup}

\input{sections/50_results}
\input{sections/60_conclusion}

\section*{Limitations}
Due to the expensive nature of our experiments, each involving continued pretraining on over 9B tokens (\S\ref{sec:dataset-details}), we could not afford to perform hyperparameter tuning, and instead took hyperparameters from prior work. It is, nevertheless, possible that careful hyperparameter tuning might yield slightly different trends from what we observed. Furthermore, because of computational constraints, we were unable to perform experiments on the largest released T5 model with 11B parameters. Though we validated our findings on two model sizes, it has been found that larger models sometimes demonstrate qualitatively different results~\citep{bigbench,lampinen-etal-2022-can,emergent}. We would be excited to see if our experiments could be reproduced at a larger model scale.

\section*{Acknowledgments}

We appreciate Victor Sanh, Albert Webson, Colin Raffel, Zaid Alyafeai, and other members of the Bigscience project who answered many of our questions when we reimplemented T0, and Yuxian Gu when we reimplemented \citet{gu-etal-2022-ppt}. We also thank Sébastien M. R. Arnold whose help was crucial for our meta-learning implementation. We are also grateful for the members at AI2 and UC Irvine, as well as the anonymous reviewers for their valuable feedback.

\bibliography{custom}
\bibliographystyle{acl_natbib}

\clearpage
\appendix
\input{sections/99_appendix}

\end{document}

%% file: math.tex
\usepackage{amsmath,amsfonts,amsthm,amssymb,bm}

\def\eqref#1{equation~\ref{#1}}

\def\1{\bm{1}}

\def\vzero{{\mathbf{0}}}

\def\vm{{\mathbf{m}}}

\def\vs{{\mathbf{s}}}

\def\mM{{\mathbf{M}}}

\DeclareMathAlphabet{\mathsfit}{\encodingdefault}{\sfdefault}{m}{sl}
\SetMathAlphabet{\mathsfit}{bold}{\encodingdefault}{\sfdefault}{bx}{n}

\newcommand{\R}{\mathbb{R}} %

\DeclareMathOperator*{\argmax}{arg\,max}

\theoremstyle{definition}

%% file: sections/10_intro.tex
\section{Introduction}

Conditioning language models (LMs) on manually-written
or learned continuous prompts allows them to solve tasks with high accuracy and minimal parameter overhead~\citep[\emph{i.a.}]{gpt3,li-liang-2021-prefix,lester-etal-2021-power}.
However, prompting performance often still lags behind traditional full finetuning.
Natural language prompts usually underperform trained models even when manually curated~\citep{gpt3,sanh2022multitask}.
Similarly, while learned prompts yield higher accuracy, they do not work as well when the training data is scarce~\citep{gu-etal-2022-ppt}, when the model is small or moderately sized~\citep{lester-etal-2021-power}, and when the tasks are difficult~\citep{he2022towards}.

To reduce the gap between prompt and full model tuning, past work has shown that
continued pretraining on data that resembles the downstream prompting setup
induces better ``promptability'', i.e., zero-shot performance with natural language (NL) prompts and few-shot performance of prompt tuning~\citep{sanh2022multitask, gu-etal-2022-ppt}.
However, in this paper, we identify several shortcomings of these methods.
First, continued pretraining on NL prompts~\citep{sanh2022multitask} sometimes causes performance degradation with prompt tuning.
Second, continued pretraining approaches that learn \emph{only} a
universal prompt initialization~\citep{gu-etal-2022-ppt,vu-etal-2022-spot} bring only marginal improvement on the P3 datasets~\citep{bach-etal-2022-promptsource}.

To further improve zero-shot and few-shot promptability, we investigate gaps in existing methods with different parameter configurations and training procedures.
First, we explore the effect of incorporating a learned continuous prompt into multi-task learning (MTL), and find it to significantly improve zero- and few-shot promptability across the board.
In addition, we explore MAML-style meta-learning~\citep{maml,reptile} as an alternative to the standard continued pretraining paradigm, but find that
it underperforms simple MTL, despite its previous success on few-shot learning tasks~\citep[\emph{i.a.}]{metasgd,gu-etal-2018-meta,qian-yu-2019-domain}.
We perform an analysis of this phenomenon and present several explanations.

Through large-scale experiments, each involving continued pretraining on over 9B tokens (\S\ref{sec:dataset-details}), we make several contributions:
(1) we thoroughly evaluate continued pretraining methods, both existing and our proposed ones, in many setups;
(2) we demonstrate that a simple continued pretraining recipe improves over existing methods by up to 31\%;
(3) we show that MAML-style meta-learning underperforms multi-task learning and provide explanations;
(4) we provide concrete recommendations to improve promptability in various use cases.

%% file: sections/20_prompting.tex
\section{Prompting} \label{sec:prompting}

We review two types of prompting that we use: natural language (NL) prompting and prompt tuning.

Traditionally, NLP tasks are solved by task-specific models that predict label $y\in\mathcal{Y}$ from input $x\in\mathcal{X}$.
We can consider LMs as functions that score any source and target text pair, $LM: \mathcal{V}^* \times \mathcal{V}^* \rightarrow \R$ with vocabulary $\mathcal{V}$.\footnote{We focus on encoder-decoder LMs based on T5~\citep{t5}.
Past work considers them to work better than decoder-only LMs for prompting~\citep{sanh2022multitask}.}
Past work found that large LMs can be repurposed to solve many tasks by casting $x,y$ into a text format using a template function $f: \mathcal{X} \cup \mathcal{Y} \rightarrow \mathcal{V}^*$ and
taking as prediction $\argmax_{y'\in\mathcal{Y}} LM(f(x), f(y'))$.

NL prompts, or instructions, are manually constructed $f(\cdot)$.
Without task-specific training, they have been successfully used to elicit predictions from LMs to perform tasks with high accuracy~\citep{gpt3,logan-iv-etal-2022-cutting}.

Sharing the motivation, prompt tuning learns a continuous prompt to condition the model.
It takes the source text embedded by the LM input embeddings, $\vs\in \R^{N\times d}$ with length $N$ and dimension $d$, and prepends learnable embeddings $E\in\R^{L\times d}$, where $L$ is a hyperparameter, to obtain a new $(L+N)$-lengthed embedded sequence. 
We consider hybrid prompt tuning, where $\vs$ is the embedding of the templatized $f(x)$,
i.e., prompt tuning is always performed \emph{in addition to} NL templates.
This has been widely adopted due to demonstrated better performance~\citep{gu-etal-2022-ppt,min-etal-2022-noisy}.
We also study a variant of prompt tuning, sometimes called prefix tuning~\citep{li-liang-2021-prefix}, where the learnable vectors are added not only to the input but all transformer layers.
See \citet{lester-etal-2021-power} and \citet{li-liang-2021-prefix} for more details on these methods.
Following the terminology of \citet{liu-etal-2022-p}, we refer to the input-level method as \textbf{shallow} prompt tuning and the layer-specific method as \textbf{deep} prompt tuning.

%% file: sections/30_improving_promptability.tex
\section{Improving Promptability} \label{sec:improving}

\rev{In this section, we describe existing methods to improve promptability and a new paradigm that combines their advantages.}

While prompt tuning sometimes performs close to full model finetuning~\citep{lester-etal-2021-power,liu-etal-2022-p}, there is often still a substantial gap, such as with limited training data~\citep{gu-etal-2022-ppt},
non-gigantic models~\citep{lester-etal-2021-power}, or challenging tasks~\citep{he2022towards}.
We therefore study ways to improve LMs' ``promptability.''
We focus on a low resource setup and consider zero-shot NL prompts and few-shot learned prompts (which, again, are in conjunction with NL prompts; \S\ref{sec:prompting}).
For the former, better promptability increases the performance when LMs face textual prompts of new tasks.
For the latter, it more effectively leverages limited training examples for higher accuracy.

We investigate if promptability can improve with a continued pretraining stage after LM pretraining (or LM adaptation for LM-adapted T5~\citep{lester-etal-2021-power}) and before task-specific finetuning.
The model is trained on a collection of tasks that have NL prompts and evaluated on unseen tasks.
\rev{The methods that we explore below differ in how the continued pretraining stage is performed. We use the notation \textbf{MTL-T\_P\_} to abbreviate those methods that are based on multi-task learning, where the blanks \textbf{\_} specify different configurations of the transformer (\textbf{T}) and the prompt (\textbf{P}) components during MTL. Architecturally, a method may continue to pretrain only the T5 model without prompt parameters, in which case we use \textcolor{noprompt}{P\xmark} to denote the lack of them; otherwise, both transformer and prompt parameters exist during MTL. We use \includegraphics[height=\fontcharht\font`\B]{fig/fire.png} and \includegraphics[height=\fontcharht\font`\B]{fig/ice.png} to denote if the corresponding component is trained or frozen in MTL, respectively. This notation describes the continued pretraining stage only: in the final finetuning stage, all methods include both the transformer and prompt components, but only the latter is updated.}

Continued pretraining has been studied in limited settings.
\citet{sanh2022multitask} proposed T0 by multi-task training a T5 model~\citep{t5} as continued pretraining. \rev{They updated T5 parameters through learning on continued pretraining tasks, not including a prompt component, and showed that this training improves zero-shot NL promptability.}
\rev{Following our nomenclature,} we refer to this paradigm as \textbf{\mFF}. %
Additionally, \citet{gu-etal-2022-ppt} \rev{employed a similar stage, incorporating and multi-task training a \emph{shallow} prompt as continued pretraining, while freezing the transformer parameters in this stage. They showed that this strategy helps few-shot promptability during finetuning.}
We refer to this paradigm as \mFT.

In this work, we study the gains of the previous two continued pretraining approaches, as well as a model that synthesizes them, \mTT, which \rev{we are the first to propose.} %
For few-shot downstream tuning, the learned prompt can act as a good initialization compared to \mFF.
In the zero-shot setup, prior work has discovered that including certain text in a prompt, such as ``Let's think step by step,'' can adjust the reasoning of LMs to yield substantially improved performance across tasks~\citep{stepbystep,anthropic}. The learned prompt here could function analogously. 
Compared to \textbf{\mFT}, on the other hand, the additional capacity brought by more updatable parameters could further boost model performance.

MAML-style meta-learning~\citep{maml} directly optimizes for the downstream updates and can outperform MTL for full model finetuning~\citep{dou-etal-2019-investigating,bansal-etal-2020-learning}.
Yet, it similarly remains unexplored for prompting.
We examine first-order MAML (\textbf{FOMAML}; \citealp{maml}), performing $T$ steps of prompt tuning in the inner loop and updating all parameters in the outer loop.
We also evaluate a version of \textbf{Reptile}~\citep{reptile} adapted for our setting that performs $T$ steps of prompt tuning followed by one step of full model tuning, and use the resulting Reptile gradient for model updates.
They have the same architecture as \textbf{\mTT} and all parameters are trainable too.
We provide a detailed description and theoretical discussion of these processes in \S\ref{sec:meta-learning}.
See the original papers for more details.

%% file: sections/40_setup.tex
\section{Experimental Setup}

We use P3, a collection of NL-templatized examples for a variety of datasets, for training and evaluation using the standard splits in \citet{sanh2022multitask}.
Not only is there no dataset overlap between training and evaluation, but no \emph{task} overlap either (e.g., sentiment vs. QA), making it challenging.
We report dataset statistics in \S\ref{sec:dataset-details}.
We perform continued pretraining for one epoch over all training datasets.
Each dataset has multiple templates, each evaluated with accuracy.
As different datasets have different numbers of answer choices and hence different baseline accuracy, we report Average Relative Gain (ARG; \citealp{ye-etal-2021-crossfit})
as a single summary metric by averaging across all templates the relative accuracy improvement over a random baseline.
We perform significance testing using bootstrap with 1,000 iterations, in each iteration randomly sampling evaluation examples and comparing the two models in question.
\S\ref{sec:per-dataset-results} reports per-dataset results.

Following \citet{sanh2022multitask}, we initialize the continued pretraining stage from T5 finetuned with an LM objective~\citep{lester-etal-2021-power}, making it more amenable to prompting.
We experiment with two sizes: T5-Large with 770M parameters and T5-XL with 3B parameters.
We retrain T0~\citep{sanh2022multitask}, i.e. \textbf{\mFF}, to eliminate confounding factors in the training procedure.
We also reproduce \citet{gu-etal-2022-ppt}'s experiment in our setup, i.e. \mFT, pretraining a \emph{shallow} prompt with other parameters frozen.
During few-shot finetuning, we train on the same 16 examples for 100 epochs. \S\ref{sec:training-details} reports additional hyperparameters.

%% file: sections/50_results.tex
\input{sections/51_results_table}

\section{Results}

Table~\ref{tab:main-results} reports our results.
From \textbf{No Cont. Pretraining}, we find that continued pretraining is crucial for prompt tuning with low resources---without it, only few-shot deep prompt tuning yields slightly above-random performance. %
These results contradict previous findings that few-shot prompt tuning works well without this stage~\citep{min-etal-2022-noisy}.
We believe this is due to the challenging nature of the P3 evaluation datasets, compared to the simple sentence classification tasks previously investigated.
This is consistent with what \citet{he2022towards} observed in the full-data setting where deep prompt tuning performs sub-optimally on difficult tasks.

\paragraph{Existing methods} for continued pretraining have their drawbacks.
In contrast to \citet{gu-etal-2022-ppt}, we found that \textbf{\mFT} with a shallow prompt does not substantially perform above random.
We attribute this to (1) their simpler evaluation tasks which, unlike ours, have decent prompt-tuned performance without continued pretraining; and (2)
their hand-designed pretraining tasks that match their evaluation tasks, while P3 conversely avoids training-evaluation task overlap, requiring generalizability.
\citet{vu-etal-2022-spot} also found \mFT to be effective, though with high resources.
We also compare with T0, i.e. \mFF, where both the official model and our reproduction suffer from degraded performance when few-shot shallow prompt tuned (compared to 0-shot), likely because the prompt added during finetuning is intrusive, and the limited gradient updates are not sufficient to recover from it.
We note that the official T0 model is not well-optimized:
even without hyperparameter tuning, our implementation is significantly better ($p<0.001$ for all).

\noindent \mTT \ \ significantly outperforms \textbf{\mFF}, the strongest existing method we examine, across all settings ($p<0.005$ for all) except for few-shot deep prompt tuning on T5-XL ($p=0.21$).
For zero-shot NL promptability, the improvement could be due to the extra model capacity, or
the multi-task trained prompt adjusting the reasoning of the LM, analogous to the text-based ``Let's think step by step'' effect~\citep{stepbystep}.
For few-shot shallow prompt tuning, unlike \textbf{\mFF}, \textbf{\mTT} does not degrade in performance, resulting in 31\% higher ARG than \mFF on T5-Large.
This is likely because of the model's familiarity with the prompt, though the limited capacity of shallow prompt tuning does not yield benefits either.
Nevertheless, with deep prompt tuning, which gives the model sufficient conditioning capacity, few-shot tuning does lead to performance increase, again outperforming \mFF.
Here, \textbf{\mTT} provides a good prompt initialization and alleviates its intrusiveness.
These results emphasize the importance of continued pretraining being aware of the downstream finetuning process.
Interestingly, however, the gap between these two models shrinks as the model size increases, no longer significant at T5-XL ($p=0.21$).
Also, notably, pretraining with a shallow prompt has better 0-shot performance than a deep prompt. This highlights that higher pretraining capacity is not always beneficial, and matches our motivation from text-based conditioning which also happens at the input level.

\paragraph{FOMAML and Reptile} surprisingly underperform \textbf{\mTT} in few-shot prompt tuning, even though they specifically optimize for this procedure and have demonstrated success in NLP for full model finetuning~\citep[\emph{i.a.}]{dou-etal-2019-investigating,bansal-etal-2020-self,bansal-etal-2021-diverse} and few-shot learning~\citep[\emph{i.a.}]{gu-etal-2018-meta,qian-yu-2019-domain,mi-etall-2019-meta}.
While \citet{ye-etal-2021-crossfit} also found FOMAML to underperform MTL, they sub-optimally only performed one inner loop update.
Here, we show that this comparison holds for more appropriate hyperparameters.
This could be due to the fewer number of gradient updates: to perform one gradient update, MTL uses one training batch, while FOMAML with $T$ inner loop steps or Reptile with $T$ prompt tuning steps use $T+1$ batches.
Not only might this be an inefficient use of training examples, but compute FLOPs too, since each inner loop/prompt tuning step involves a full forward-backward pass.
We attempt using a $T+1$ times smaller meta batch size (see \S\ref{sec:meta-learning} for more detail) to pretrain a deep T5-Large-sized Reptile. When prompt-tuned, it achieves 22.8 ARG, which is even lower, possibly due to higher gradient estimation noise.
Alternatively, other factors could affect the performance of meta-learning.
It is, for example, well-known that MAML-style meta-learning can be unstable and sensitive to architectures and hyperparameters~\citep{antoniou2018how}.
This instability is likely amplified by our large heterogeneous multi-task setup and our inability to afford hyperparameter search.
Furthermore, its theoretical foundation has mostly only been examined through simple optimizers, predominantly SGD~\citep{maml,reptile}.
How it interacts with optimizers more common in modern NLP, such as Adafactor (which we use), remains to be explored.

\paragraph{Recommendations.}
Based on our findings, we recommend practitioners to always incorporate a prompt during continued pretraining and to train the entire model. Without downstream task-specific tuning, such as when there is no training data or sufficient compute, a shallow prompt yields better accuracy. When few-shot task-specific prompt tuning is affordable, continued pretraining with a deep prompt enables the best performance.

%% file: sections/51_results_table.tex
\begin{table*}[t!]
    \centering
    \begin{tabular}{@{\hspace{0pt}}l@{\hspace{5pt}}c@{\hspace{4pt}}c@{\hspace{5pt}}c@{\hspace{4pt}}c@{\hspace{5pt}}c@{\hspace{4pt}}c@{\hspace{5pt}}c@{\hspace{4pt}}c@{\hspace{0pt}}}
        \toprule
        & \multicolumn{4}{c}{\textbf{T5-Large (770M)}} & \multicolumn{4}{c}{\textbf{T5-XL (3B)}} \\
        \cmidrule(lr){2-5} \cmidrule(lr){6-9}
        & \multicolumn{2}{c}{\textbf{Shallow}} & \multicolumn{2}{c}{\textbf{Deep}} & \multicolumn{2}{c}{\textbf{Shallow}} & \multicolumn{2}{c}{\textbf{Deep}} \\
        \cmidrule(lr){2-3} \cmidrule(lr){4-5}
        \cmidrule(lr){6-7}
        \cmidrule(lr){8-9}
        & \textbf{0-shot} & \textbf{16-shot} & \textbf{0-shot} & \textbf{16-shot} & \textbf{0-shot} & \textbf{16-shot} & \textbf{0-shot} & \textbf{16-shot} \\
        \midrule
        \textbf{No Cont. Pretraining}{\small~(LM-adapted T5)} & \phantom{$^*$}--1.9$^*$ & --1.5 & \phantom{$^*$}--1.9$^*$ & \phantom{0}2.3 & \phantom{$^*$}--1.5$^*$ & --1.5 & \phantom{$^*$}--1.5$^*$ & \phantom{0}3.7 \\
        \midrule
        \\[-5.8mm]
        \emph{Previous methods} \\
        \textbf{\mFT}{\small~(Ours à la \citet{gu-etal-2022-ppt})} & \phantom{0}2.8 & \phantom{0}2.9 & --- & --- & --1.7 & --1.6 & --- & --- \\
        \textbf{\mFF}{\small~\citep{sanh2022multitask}} & --- & --- & --- & --- & \phantom{$^*$}25.2$^*$ & 19.1 & \phantom{$^*$}25.2$^*$ & 33.8 \\
        \textbf{\mFF}{\small~(Our reproduction)} & \phantom{$^*$}26.7$^*$ & 23.1 & \phantom{$^*$}26.7$^*$ & 32.6 & \phantom{$^*$}32.4$^*$ & 30.0 & \phantom{$^*$}32.4$^*$ & 42.9 \\
        \midrule
        \\[-5.8mm]
        \emph{Our methods} \\
        \textbf{\mTT} & \textbf{30.2} & \textbf{30.2} & \textbf{28.7} & \textbf{37.4} & \textbf{33.5} & \textbf{33.5} & \textbf{33.3} & \textbf{43.2} \\
        \textbf{FOMAML} & 21.8 & 21.8 & 20.8 & 32.8 & 27.7 & 27.7 & 24.5 & 40.0 \\
        \textbf{Reptile} & 18.4 & 18.4 & 24.9 & 33.0 & 23.1 & 23.2 & 26.3 & 39.7 \\
        \bottomrule
    \end{tabular}
    \caption{
        \label{tab:main-results}
        Average Relative Gain (ARG) on the P3 evaluation datasets. Random performance is 0.0 ARG.
        Each row in \emph{Our methods} represents four pretrained models, for Large/XL $\times$ Shallow/Deep, while \textbf{\mFF} shares the pretrained model for Shallow and Deep as it has no pretrained prompt.
        We \textbf{bold} the highest number in each column.
        $^*$: These models have no prompt and hence no Shallow/Deep distinction in 0-shot experiments.
    }
\end{table*}

%% file: sections/60_conclusion.tex
\section{Conclusion}

We demonstrated that the simple recipe of continued pretraining with a prompt significantly improves zero-shot NL promptability and few-shot learned promptability. MAML-based meta-learning, on the other hand, obtains worse performance, for which we provided several explanations. Nonetheless, we believe future efforts to leverage their conceptual advantage could be fruitful, perhaps aided by our observations. We also hope to study the effect of continued pretraining with other parameter injection methods~\citep{adapters,hu2022lora,tfew}.

%% file: sections/99_appendix.tex
\setlength{\algomargin}{0.7em}
\makeatletter
\patchcmd{\@algocf@start}%
  {-1.5em}%
  {-0.5em}%
  {}{}%
\makeatother
\begin{algorithm}[t]
\SetKwInput{KwInput}{Input}
\SetKw{KwInit}{Initialize}
\SetKw{KwIn}{in}
\DontPrintSemicolon
\KwInput{Number of inner loop steps $T$, meta batch size $B$}
\KwInit LM-adapted T5 parameters $\phi^{\text{orig}}$\;

global\_grad = $\vzero$\;
\For{b $\leftarrow$ 1, $\cdots$, $B$} {
    $\phi$ = clone($\phi^{\text{orig}}$)\;
    \For{t $\leftarrow$ 1, $\cdots$, $T$} {
        data = next\_batch() \tcp*{support}
        grad = forward\_backward($\phi$, data)\;
        update($\phi$, grad, prompt\_only=True)\;
    }
    data = next\_batch() \tcp*{query}
    grad = forward\_backward($\phi$, data)\;
    global\_grad += grad\;
}
global\_grad = global\_grad / $B$\;
update($\phi^{\text{orig}}$, global\_grad, prompt\_only=False)\;
\caption{The FOMAML algorithm for prompt tuning.}
\label{alg:fomaml}
\end{algorithm}

\begin{algorithm}[t]
\SetKwInput{KwInput}{Input}
\SetKw{KwInit}{Initialize}
\SetKw{KwIn}{in}
\DontPrintSemicolon
\KwInput{Number of inner loop steps $T$, meta batch size $B$, inner loop learning rate $\alpha$}
\KwInit LM-adapted T5 parameters $\phi^{\text{orig}}$\;

global\_grad = $\vzero$\;
\For{b $\leftarrow$ 1, $\cdots$, $B$} {
    $\phi$ = clone($\phi^{\text{orig}}$)\;
    \For{t $\leftarrow$ 1, $\cdots$, $T$} {
        data = next\_batch() \tcp*{support}
        grad = forward\_backward($\phi$, data)\;
        update($\phi$, grad, prompt\_only=True)\;
    }
    data = next\_batch() \tcp*{query}
    grad = forward\_backward($\phi$, data)\;
    update($\phi$, grad, prompt\_only=False)\;
    global\_grad --= $\phi$\;
}
global\_grad = global\_grad / $(\alpha B)$ + $\phi^{\text{orig}}$\;
update($\phi^{\text{orig}}$, global\_grad, prompt\_only=False)\;
\caption{The Reptile algorithm for prompt tuning.}
\label{alg:reptile}
\end{algorithm}

\section{Dataset Details} \label{sec:dataset-details}

We use P3 as our training and evaluation datasets~\citep{bach-etal-2022-promptsource}. It contains 35 datasets grouped into 8 tasks: Multiple-Choice QA, Extractive QA, Closed-Book QA, Sentiment, Topic Classification, Structure-To-Text, Summarization, and Paraphrase Identification. Examples in each dataset are templatized using multiple human-written templates. Across the 35 datasets, there are a total of 313 templates. For continued pretraining, we follow \citet{sanh2022multitask} and only use the training split of each dataset. Four tasks are held out for evaluation in P3: Sentence Completion, Natural Language Inference, Coreference Resolution, and Word Sense Disambiguation. They consist of 11 evaluation datasets (considering the three splits of ANLI as separate datasets) and 116 templates in total. We use the training split of each dataset for few-shot experiments, and, following \citet{sanh2022multitask}, evaluate on the validation splits. The only exception is StoryCloze which does not have a training split, so we use its validation split for training and evaluate on its test split. Unlike T0, we do not evaluate on the BIG-Bench datasets~\citep{bigbench} as they had not stabilized as a collection of datasets at the time of this work. All the prompts in P3 are collected in English.

To make training more efficient, we right-truncate all source sequences to 768 tokens and target sequences to 192 tokens. For the continued pretraining stage, this affects 2\% of all training examples, and among the 313 templates, 24 have more than 1\% examples truncated. Also, following \citet{sanh2022multitask}, we cap all datasets to have a maximum of 500k examples. This results in 31.6M training examples across all datasets and templates, totaling 5.7B tokens on the source side and 3.5B tokens on the target side.

\section{Meta-Learning Details} \label{sec:meta-learning}

In this section, we elaborate on our meta-learning training procedures. Algorithm~\ref{alg:fomaml} contains pseudo-code for our first-order MAML (FOMAML) procedure. In the inner loop, we perform $T$ steps of prompt tuning on a cloned model using support data. In the outer loop, we use query data to evaluate the prompt-tuned model and compute gradients. We use the first-order approximation where the gradient is not taken with respect to the entire prompt tuning process but only the forward pass with query data because it is computationally more tractable, and past work has shown that this first-order approximation does not hurt performance much, if at all~\citep{maml,dou-etal-2019-investigating}. Theoretically, to perfectly simulate the downstream prompt tuning procedure, we should use the same batch of support data for the $T$ steps of update. Nevertheless, this would traverse the training data much more slowly, so we use different support batches. Our theoretical analysis through the perspective of Reptile below also justifies this.

In preliminary experiments, we found a na\"{i}ve adoption of Reptile~\citep{reptile} to yield subpar performance. As there is no inner- and outer-loop distinction in Reptile, doing prompt tuning leads to only the prompt parameter being updated throughout the entire continued pretraining stage, likely causing the performance degradation. This effect is also seen in our multi-task learning setup with the \textbf{\mFT} model. Thus, we propose to adapt Reptile to better suit prompt tuning, which we illustrate in Algorithm~\ref{alg:reptile}. It is similar to FOMAML, but instead of considering the outer loop's gradient as the meta-learning gradient, it uses $\frac{1}{\alpha B}\sum_{b=1}^{B} (\phi^\text{orig} - \phi^b)$ where $b$ is the cloned model's final parameter for meta-batch $b$.

Now we theoretically justify our proposed Reptile version, mostly following the original proof structure in \citet{reptile}, in the context of prompt tuning. We can think of the downstream finetuning stage as starting from the initial model parameters $\phi_0$ and performing $T$ steps of prompt tuning on the same batch of training data which produces a loss function $L_{\text{train}}(\phi)$ for some model parameters $\phi$.
Then this model is evaluated on some test data that similarly produces a loss function $L_{\text{test}}(\phi_T)$ for the final trained model $\phi_T$. Let us first abbreviate some gradients and Hessians: 
\begin{align*}
    g_i&=L'_{\text{train}}(\phi_i) = \frac{\partial}{\partial \phi_i}L_{\text{train}}(\phi_i) \\
    H_i&=L''_{\text{train}}(\phi_i) \\
    \overline{g}&=L'_{\text{test}}(\phi_0) \\
    \overline{H}&=L''_{\text{test}}(\phi_0)
\end{align*}

Then we can write each step of the prompt tuning process as an update function:
\begin{align*}
    U(\phi) &= \phi - \alpha \vm \circ L'_{\text{train}}(\phi) \\
    U'(\phi) &= I - \alpha \mM \circ L''_{\text{train}}(\phi)
\end{align*}
where $\alpha$ is the learning rate and $\vm$ and $\mM$ are boolean masks that contain 1 for the prompt parameters. $\circ$ indicates element-wise multiplication, which we prescribe to take the highest precedence in the equations below.

With $T$ iterations of $U$, we have:
\begin{align} \label{eq:reptile-update}
    \phi_T = \phi_0 - \alpha \vm \circ \sum_{j=0}^{T-1} g_j
\end{align}

Plugging in Equation~\ref{eq:reptile-update}, by Taylor's theorem:
\begin{align}
\begin{split} \label{eq:reptile-gi}
    g_i &= L'_{\text{train}}(\phi_i) \\
    &= L'_{\text{train}}(\phi_0) + L''_{\text{train}}(\phi_0)(\phi_i-\phi_0)+\mathcal{O}(\alpha^2) \\
    &= g_0 + H_0 (\phi_i - \phi_0) +\mathcal{O}(\alpha^2) \\
    &= g_0 - \alpha H_0 \vm \circ \sum_{j=0}^{i-1} g_j +\mathcal{O}(\alpha^2) \\
    &= g_0 - \alpha i H_0 \vm \circ g_0 + \mathcal{O}(\alpha^2)
\end{split}
\end{align}
where the last step can be seen by induction, iteratively applying the second-to-last line.

With a similar process, we can derive:
\begin{align} \label{eq:reptile-hi}
    H_i = H_0 + \mathcal{O}(\alpha)
\end{align}

The FOMAML gradient is the same as $L'_{\text{test}}(\phi_T)$. Plugging in Equation~\ref{eq:reptile-gi} but sweeping its non-leading terms into $\mathcal{O}(\alpha^2)$:
\begin{align*}
    g_{\text{FOMAML}} &= L'_{\text{test}}(\phi_T) \\
    &= L'_{\text{test}}(\phi_0) + L''_{\text{test}}(\phi_0)(\phi_T-\phi_0) \\
    & \quad +\mathcal{O}(\alpha^2) \\
    &= \overline{g} + \overline{H} (\phi_T - \phi_0) + \mathcal{O}(\alpha^2) \\
    &= \overline{g} - \alpha \overline{H} \vm \circ \sum_{j=0}^{T-1} g_j + \mathcal{O}(\alpha^2) \\
    &= \overline{g} - \alpha T \overline{H} \vm \circ g_0 + \mathcal{O}(\alpha^2)
\end{align*}

The full MAML gradient takes the derivative throughout the entire prompt tuning process. Plugging in Equation~\ref{eq:reptile-hi} and $g_{\text{FOMAML}} = L'_{\text{test}}(\phi_T)$ and sweeping terms into $\mathcal{O}(\alpha^2)$ when possible:

\vspace{-1.5em}
{\fontsize{10}{12} \selectfont
\begin{align*}
    g_{\text{MAML}} &= \frac{\partial}{\partial\phi_0} L_{\text{test}}(\phi_T) \\
    &= \frac{\partial}{\partial\phi_0} L_{\text{test}}(U(U(\cdots U(\phi_0)))) \\
    &= U'(\phi_0)U'(\phi_1)\cdots U'(\phi_{T-1})L'_{\text{test}}(\phi_T) \\
    &= \left(\prod_{j=0}^{T-1}\left(I - \alpha \mM \circ L''_{\text{train}}(\phi_j)\right)\right) L'_{\text{test}}(\phi_T) \\
    &= \left(\prod_{j=0}^{T-1}\left(I - \alpha \mM \circ H_j\right)\right) L'_{\text{test}}(\phi_T) \\
    &= \left(I - \alpha \mM \circ \sum_{j=0}^{T-1} H_j\right) L'_{\text{test}}(\phi_T) \\
    & \quad + \mathcal{O}(\alpha^2)\\
    &= \left(I - \alpha \mM \circ \sum_{j=0}^{T-1} H_j\right) \left(\overline{g} - \alpha T \overline{H} \vm \circ g_0\right) \\
    & \quad + \mathcal{O}(\alpha^2) \\
    &= \left(I - \alpha T \mM \circ H_0\right) \left(\overline{g} - \alpha T \overline{H} \vm \circ g_0\right) \\
    & \quad + \mathcal{O}(\alpha^2) \\
    &= \overline{g} - \alpha T \mM \circ H_0 \overline{g} - \alpha T \overline{H} \vm \circ g_0 \\
    & \quad + \mathcal{O}(\alpha^2)
\end{align*}
}

The Reptile gradient, in our adaptation, takes the prompt's gradient during the $T$ steps and the entire model's gradient for one step. Taking the Reptile gradient from Algorithm~\ref{alg:reptile} and using $\phi_{T+1}$ to represent the parameters after the outer loop full-finetuning update:
\begin{align*}
    g_{\text{Reptile}} &= \frac{1}{\alpha} (\phi_0 - \phi_{T+1}) \\
    &= \vm \circ \sum_{j=0}^{T-1}g_j + g_{\text{FOMAML}} \\
    &= \vm \circ \sum_{j=0}^{T-1} \left(g_0 - \alpha j H_0 \vm \circ g_0\right) \\
    & \qquad + g_{\text{FOMAML}} + \mathcal{O}(\alpha^2) \\
    &= T\vm \circ g_0 - \alpha\frac{T(T-1)}{2}\vm \circ (H_0 \vm \circ g_0) \\
    & \qquad + g_{\text{FOMAML}} + \mathcal{O}(\alpha^2) \\
    &= \overline{g} + T\vm \circ g_0 \\
    & \qquad - \alpha\frac{T(T-1)}{2}\vm \circ (H_0 \vm \circ g_0) \\
    & \qquad - \alpha T \overline{H} \vm \circ g_0 + \mathcal{O}(\alpha^2)
\end{align*}

We can see that all three meta-learning gradients have a similar effect: they only contain a mixture of lone gradients terms ($\overline{g}, g_0$), which act as a pure multi-task learning objective, and terms that are Hessian times gradient, which \citet{reptile} termed ``AvgGradInner'' and showed to encourage the expected similarity between different data batches, improving generalization.

Back to our use of different data batches in FOMAML's inner loop and Reptile's prompt tuning steps. If the inner loop uses the same support (i.e., training) data, as in the derivation above, the ``AvgGradInner'' terms become somewhat degenerate, with the same term scaled $T$ or $\frac{T(T-1)}{2}$ times. With different inner loop batches, on the other hand, there would be more diverse Hessian-gradient interactions between different batches of data and hence encouraging generalization between more tasks.

\section{Training Details} \label{sec:training-details}

Due to the expensiveness of our experiments, we did not perform any hyperparameter tuning. For all continued pretraining runs, we follow \citet{t5} and \citet{sanh2022multitask} and use Adafactor~\citep{adafactor} with a 0.001 learning rate. We use a batch size of 4,096 which we calculated to be close to what \citet{sanh2022multitask} used.\footnote{They used example packing by putting multiple examples in one sequence to better suit TPUs, which we didn't use, so the batch sizes are not perfectly comparable.} We clip gradients to unit norm. For shallow prompt tuning, we follow \citet{min-etal-2022-noisy} and use $L=20$ prompt tokens, each with the same dimension as the word embedding size, on the source side only. For deep prompt tuning, we similarly use 20 hidden vectors that are prepended in every transformer layer, on both the source and target side for added capacity. For meta-learning, we use a batch size of 16, simulating our 16-shot evaluation (see below), and a meta batch size of 128. We perform 7 steps of inner loop updates (FOMAML) / prompt tuning (Reptile), following \citet{bansal-etal-2020-self} and \citet{bansal-etal-2021-diverse}, and similarly using Adafactor with learning rate 0.001. All continued pretraining experiments run for one epoch over the training datasets with no checkpoint selection. In few-shot finetuning, we train on one batch of 16 randomly selected examples for 100 epochs (the same batch throughout training), following \citet{min-etal-2022-noisy}. Like \citet{min-etal-2022-noisy}, we do not manually balance the label distribution in these examples, unlike in prior work~\citep{gao-etal-2021-making,logan-iv-etal-2022-cutting}.

We perform all experiments on 80GB A100 GPUs. Each continued pretraining run takes four (sometimes eight) of them. The largest MTL model takes 10 days to pretrain with four GPUs, while the largest meta-learning model takes 14 days.

\section{Per-Dataset Results} \label{sec:per-dataset-results}

\begin{figure*}[t!]
	\centering
	\includegraphics[width=\textwidth]{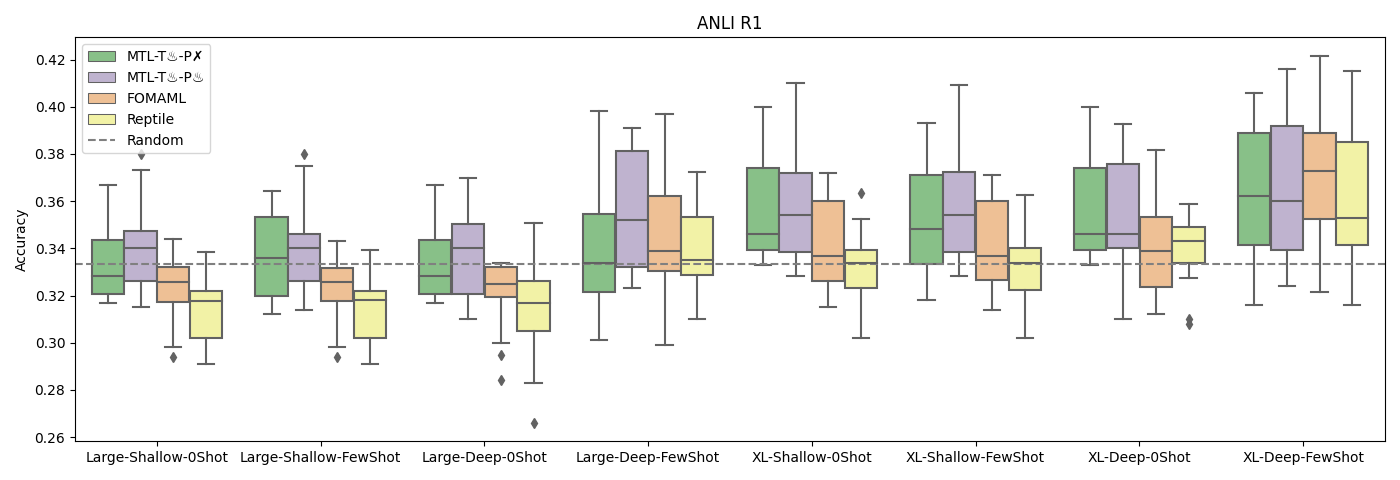}
	\includegraphics[width=\textwidth]{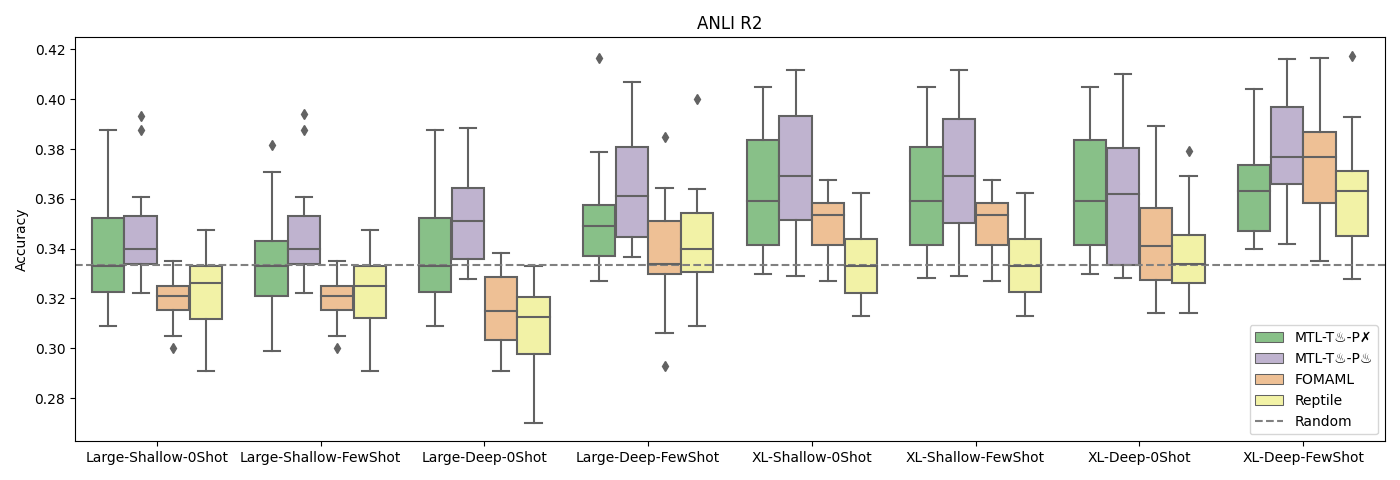}
    \includegraphics[width=\textwidth]{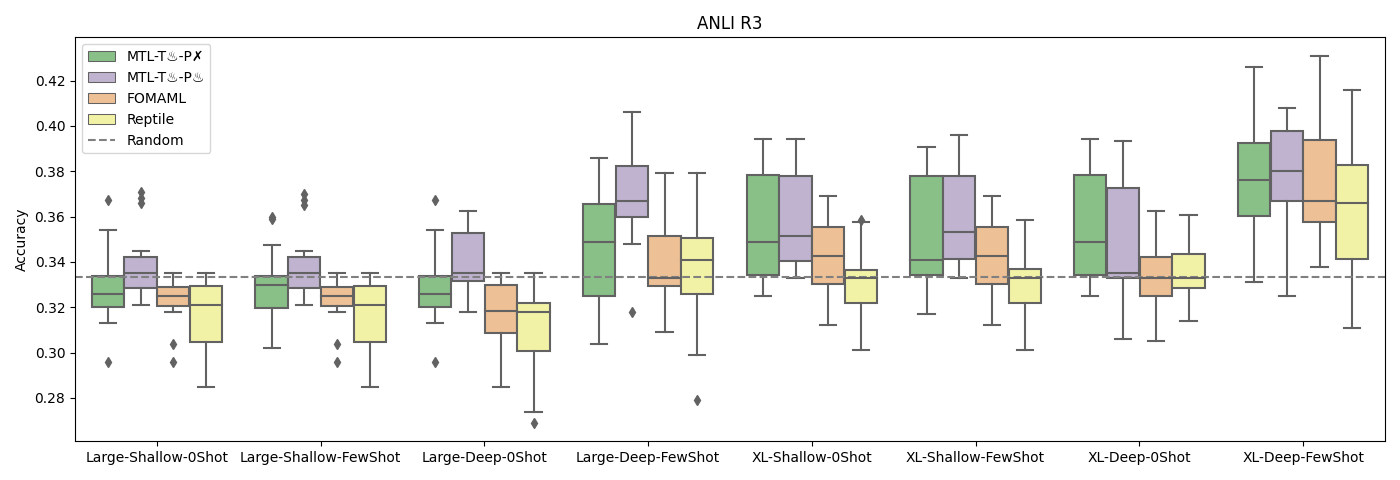}
    \includegraphics[width=\textwidth]{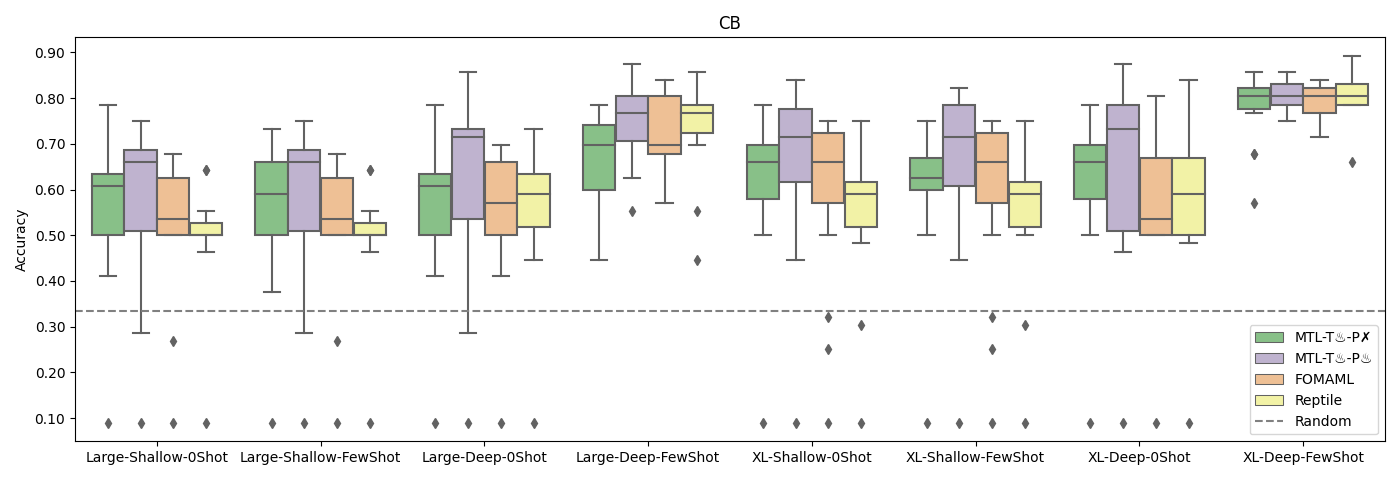}
	\caption{Per-dataset accuracy of \mFF (our reproduction), \mTT, \textbf{FOMAML}, \textbf{Reptile}, and the random baseline (part 1).}
	\label{fig:per-dataset-p1}
\end{figure*}

\begin{figure*}[t!]
	\centering
	\includegraphics[width=\textwidth]{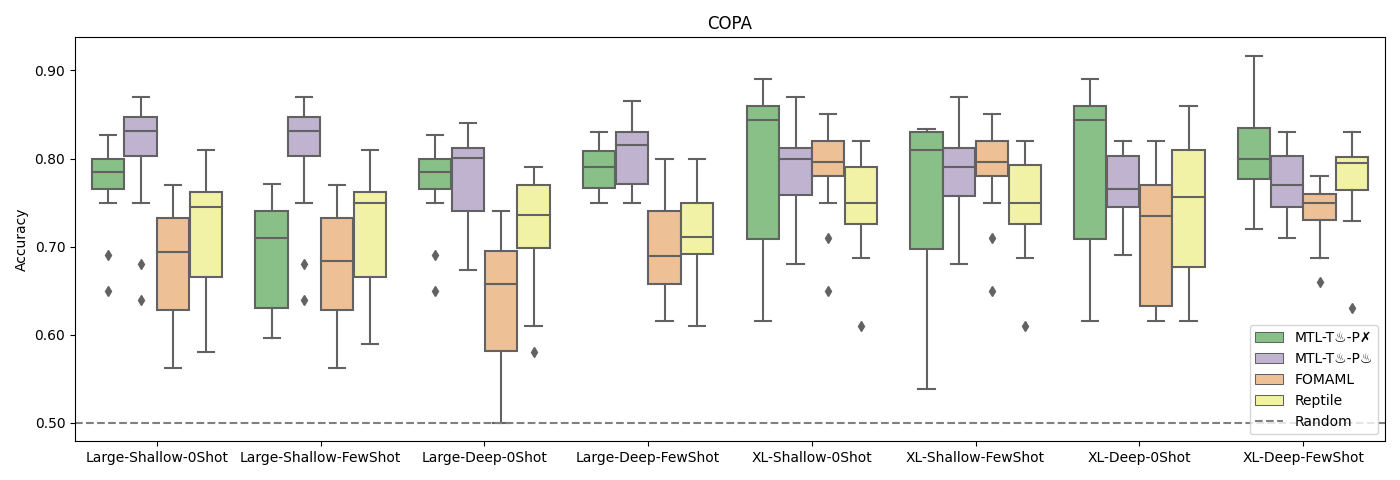}
	\includegraphics[width=\textwidth]{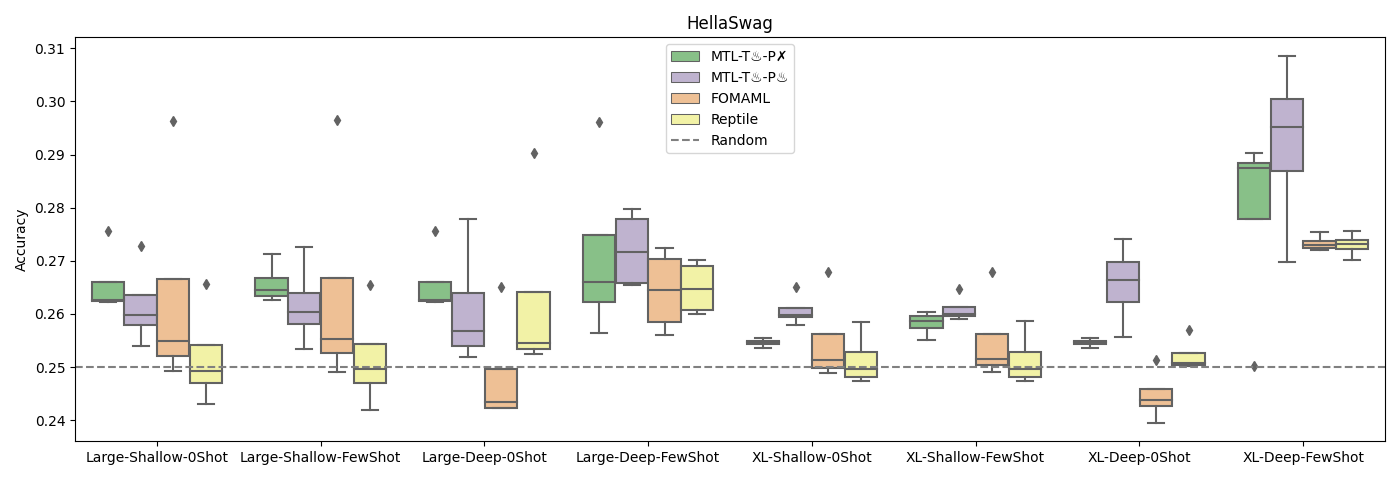}
    \includegraphics[width=\textwidth]{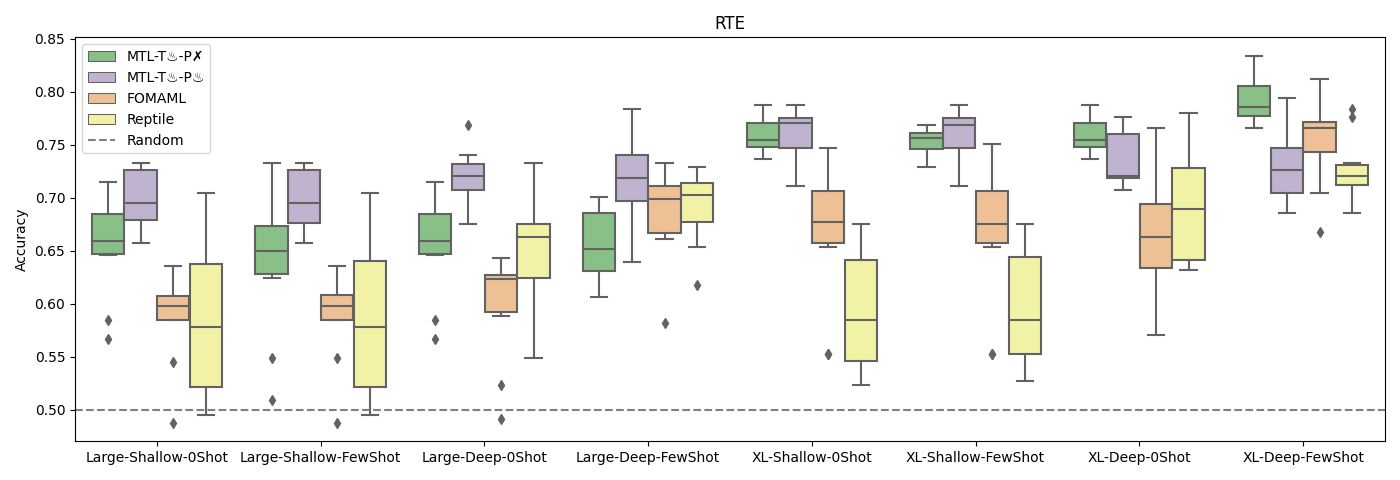}
    \includegraphics[width=\textwidth]{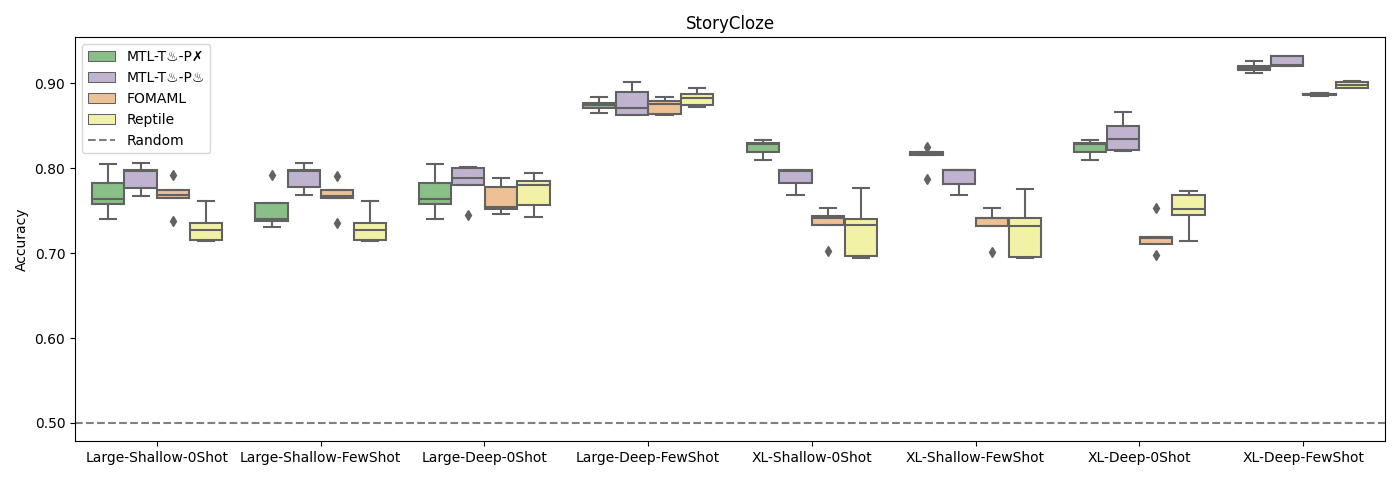}
	\caption{Per-dataset accuracy of \mFF (our reproduction), \mTT, \textbf{FOMAML}, \textbf{Reptile}, and the random baseline (part 2).}
	\label{fig:per-dataset-p2}
\end{figure*}

\begin{figure*}[t!]
	\centering
	\includegraphics[width=\textwidth]{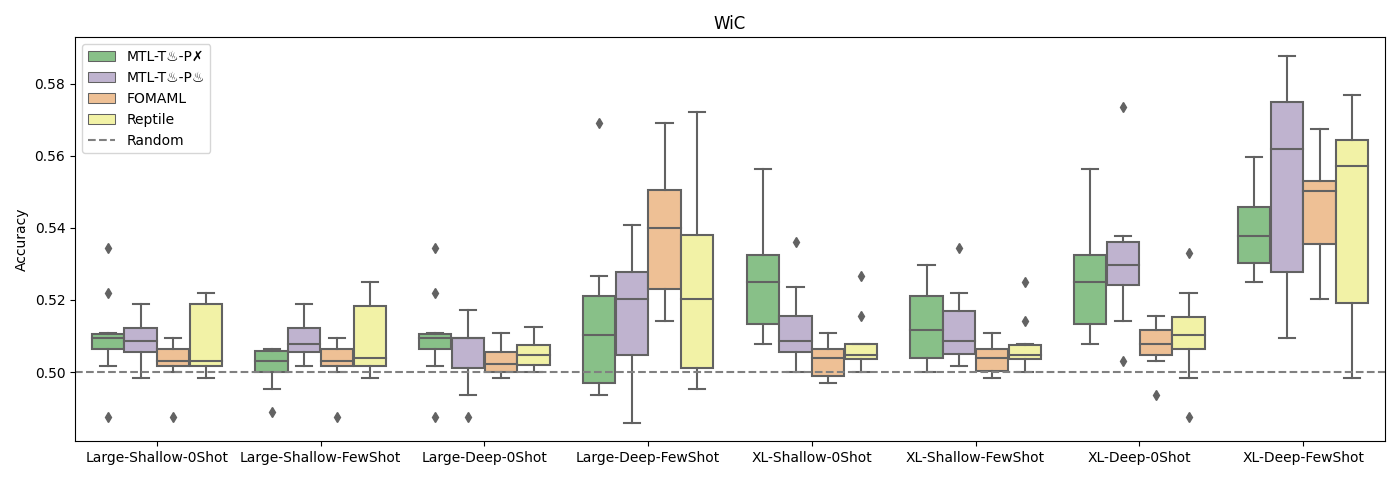}
	\includegraphics[width=\textwidth]{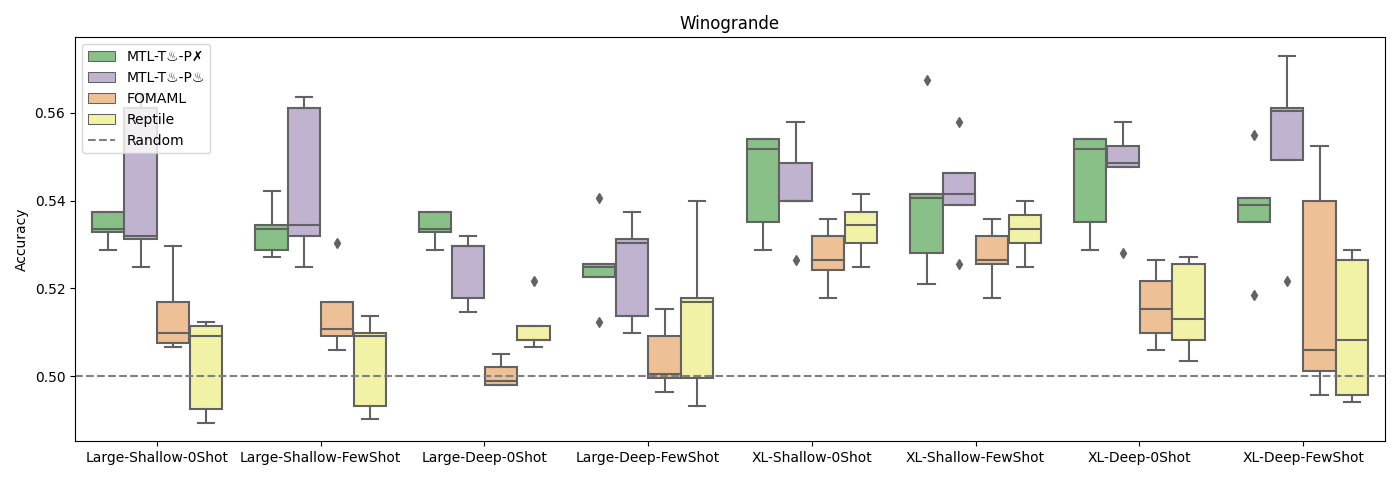}
    \includegraphics[width=\textwidth]{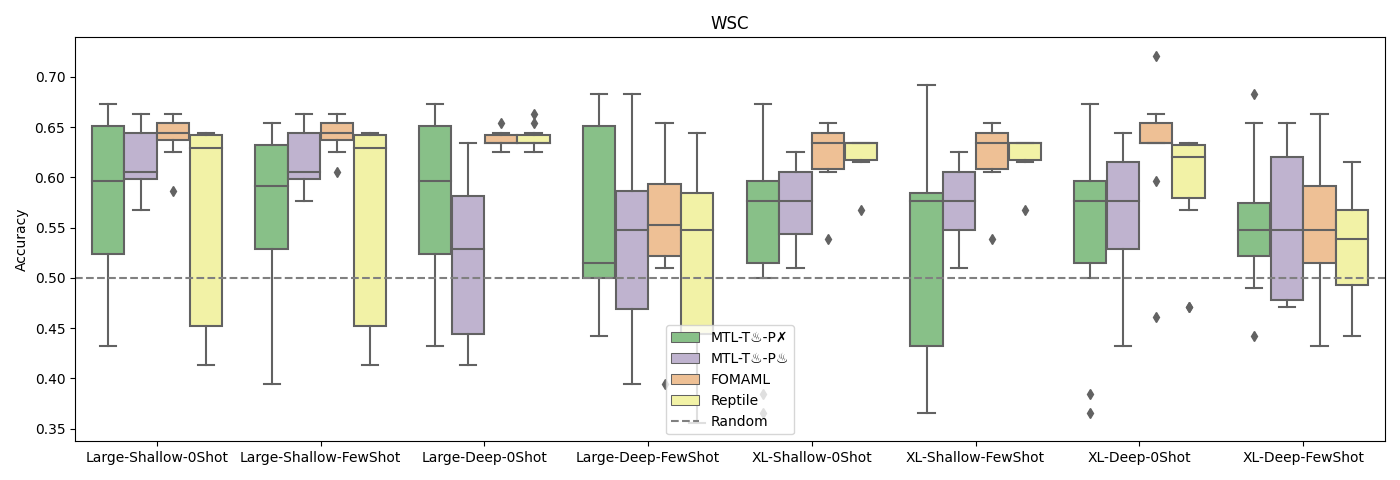}
	\caption{Per-dataset accuracy of \mFF (our reproduction), \mTT, \textbf{FOMAML}, \textbf{Reptile}, and the random baseline (part 3).}
	\label{fig:per-dataset-p3}
\end{figure*}

In Figures~\ref{fig:per-dataset-p1} to \ref{fig:per-dataset-p3}, we compare the per-dataset \emph{accuracy} of \mFF (our reproduction), \mTT, \textbf{FOMAML}, and \textbf{Reptile}. We omit \mFT due to its near-random performance.

